\documentclass[11pt,onecolumn,letterpaper]{article}

\usepackage[letterpaper,left=0.70in,right=0.70in,top=0.78in,bottom=0.78in,headheight=24pt,headsep=16pt,footskip=25pt]{geometry}
\usepackage[T1]{fontenc}
\usepackage[lining]{ebgaramond}
\usepackage{fix-cm}
\usepackage{graphicx,amsmath,amssymb,booktabs}
\usepackage{ebgaramond-maths}
\usepackage[dvipsnames]{xcolor}
\usepackage[numbers,sort&compress]{natbib}
\usepackage{caption,subcaption,enumitem,etoolbox}
\usepackage{fancyhdr,titlesec,placeins}
\usepackage[most]{tcolorbox}

\usepackage[stretch=30,shrink=30]{microtype}

\newcommand{\geoflip}{\textit{GeoFlip}}
\definecolor{geometrygreen}{HTML}{397E76}
\definecolor{geometryorange}{HTML}{B87747}

\usepackage{colortbl}
\usepackage{tabularx}
\usepackage{pifont}
\definecolor{comparisonlavender}{HTML}{F2EEF9}
\definecolor{comparisonpositive}{HTML}{16A05D}
\definecolor{comparisonnegative}{HTML}{D43838}
\definecolor{comparisonpartial}{HTML}{9A5B08}
\newcommand{\gfyes}{\textcolor{comparisonpositive}{\raisebox{-0.6pt}{\fontsize{12}{12}\selectfont\ding{52}}}}
\newcommand{\gfno}{\textcolor{comparisonnegative}{\raisebox{-0.6pt}{\fontsize{12}{12}\selectfont\ding{56}}}}
\newcommand{\gfpartial}{\textcolor{comparisonpartial}{\raisebox{-0.3pt}{\fontsize{12}{12}\selectfont$\blacktriangle$}}}

\usepackage{tikz}

\definecolor{reportink}{HTML}{252335}
\definecolor{reportaccent}{HTML}{655080}
\definecolor{reportrule}{HTML}{DCD4E7}
\definecolor{reportfront}{HTML}{F6F8FB}
\usepackage[breaklinks,colorlinks,linkcolor=blue!55!black,citecolor=blue!55!black,urlcolor=blue!60!black]{hyperref}
\usepackage[capitalize,nameinlink]{cleveref}
\crefname{figure}{Fig.}{Figs.}
\Crefname{figure}{Figure}{Figures}
\hypersetup{pdftitle={GeoFlip Preview: Benchmarking Geometry-Sensitive Physics in Video Generation},pdfauthor={Yuqi Li; Junming Liu; Bowen Hu; Ruize Niu; Huiran Duan; Ming-Hsuan Yang; Yingli Tian; Hao Wu}}
\AtBeginEnvironment{figure}{\captionsetup{width=0.86\textwidth}}
\AtBeginEnvironment{table}{\captionsetup{width=0.86\textwidth}}
\let\reportincludegraphics\includegraphics
\renewcommand{\includegraphics}[2][]{%
 \ifstrequal{#2}{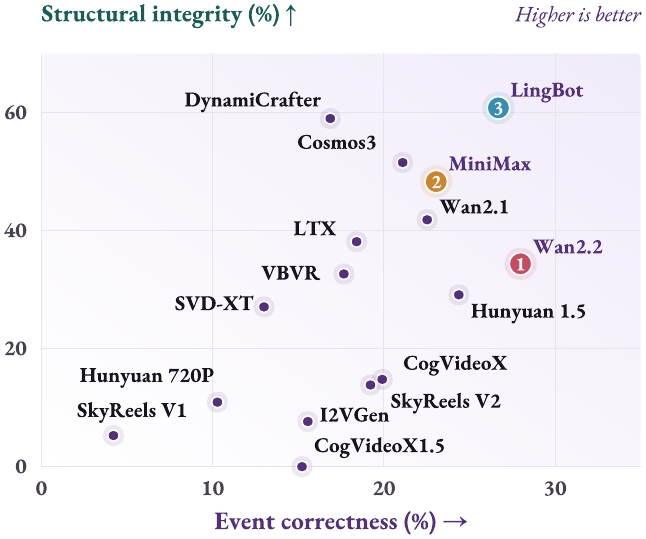}{\reportincludegraphics[width=0.60\linewidth]{#2}}{%
 \ifstrequal{#2}{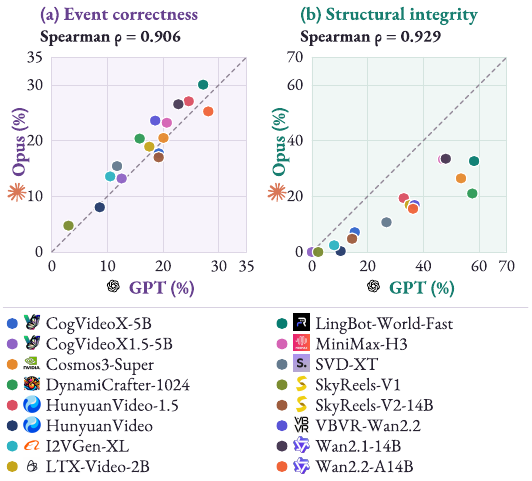}{\reportincludegraphics[width=0.78\linewidth]{#2}}{%
 \ifstrequal{#2}{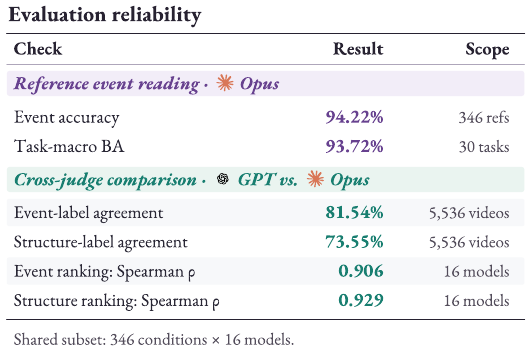}{\reportincludegraphics[width=0.60\linewidth]{#2}}{%
 \reportincludegraphics[#1]{#2}}}}}
\titleformat{\section}{\Large\bfseries\color{reportink}}{\thesection}{0.65em}{}
\titleformat{\subsection}{\large\bfseries\color{reportink}}{\thesubsection}{0.65em}{}
\titlespacing*{\section}{0pt}{16pt plus 3pt minus 2pt}{7pt}
\titlespacing*{\subsection}{0pt}{11pt plus 2pt minus 1pt}{5pt}
\renewcommand{\headrulewidth}{0.4pt}
\renewcommand{\headrule}{\hbox to\headwidth{\color{reportrule}\leaders\hrule height\headrulewidth\hfill}}
\fancypagestyle{firstpage}{\fancyhf{}\fancyfoot[C]{\small\thepage}\renewcommand{\headrulewidth}{0pt}}
\makeatletter
\newenvironment{reportfrontmatter}{%
 \thispagestyle{firstpage}%
 \begin{tcolorbox}[enhanced,colback=reportfront,frame hidden,boxrule=0pt,
 arc=11pt,left=16pt,right=16pt,top=16pt,bottom=15pt,
 boxsep=0pt,before skip=0pt,after skip=15pt]
 {\raggedright\fontsize{24}{27}\selectfont\bfseries\@title\par}
 \vspace{9pt}
 {\fontsize{11.8}{14.2}\selectfont\bfseries\reportauthors\par}
 \vspace{8pt}
 {\fontsize{10.8}{13.3}\selectfont\reportaffiliations\par}
 \vspace{3pt}
 {\small\reportauthornotes\par}
}{\end{tcolorbox}}
\renewenvironment{abstract}{\par\vspace{11pt}\begingroup\normalsize\setlength{\parindent}{0pt}\ignorespaces}{\par\endgroup}
\makeatother
\newcommand{\reportshorttitle}{\textit{GeoFlip Preview}}
\newcommand{\reporttype}{Technical Report}

\title{\geoflip{} Preview: Benchmarking Geometry-Sensitive Physics in Video Generation}
\author{
Yuqi Li\textsuperscript{1,*}\quad
Junming Liu\textsuperscript{2,*}\quad
Bowen Hu\textsuperscript{3}\quad
Ruize Niu\textsuperscript{4}\\
Huiran Duan\textsuperscript{1}\quad
Ming-Hsuan Yang\textsuperscript{5}\quad
Yingli Tian\textsuperscript{1}\quad
Hao Wu\textsuperscript{6,\ensuremath{\dagger}}\\[4pt]
\textsuperscript{1}City University of New York, USA\\
\textsuperscript{2}University of Melbourne, Australia\\
\textsuperscript{3}Sun Yat-sen University, China\\
\textsuperscript{4}The Chinese University of Hong Kong, Hong Kong, China\\
\textsuperscript{5}University of California, Merced, USA\\
\textsuperscript{6}Carnegie Mellon University, USA\\[4pt]
{\small \textsuperscript{*}Equal contribution.
\quad \textsuperscript{\ensuremath{\dagger}}Corresponding author.}\\
{\ttfamily\small hw5@andrew.cmu.edu}
}

\newcommand{\reportauthors}{%
Yuqi Li\textsuperscript{1,*}, Junming Liu\textsuperscript{2,*}, Bowen Hu\textsuperscript{3}, Ruize Niu\textsuperscript{4},\\[2pt]
Huiran Duan\textsuperscript{1}, Ming-Hsuan Yang\textsuperscript{5}, Yingli Tian\textsuperscript{1}, Hao Wu\textsuperscript{6,\ensuremath{\dagger}}}
\newcommand{\reportaffiliations}{%
\begin{tabular}{@{}l@{}}
\raisebox{-4pt}{\includegraphics[height=16pt]{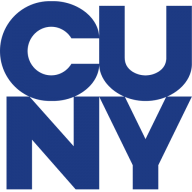}}\hspace{5pt}\textsuperscript{1}City University of New York, USA\\[3pt]
\raisebox{-4pt}{\includegraphics[height=16pt]{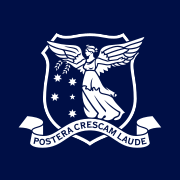}}\hspace{5pt}\textsuperscript{2}University of Melbourne, Australia\\[3pt]
\raisebox{-4pt}{\includegraphics[height=16pt]{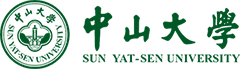}}\hspace{5pt}\textsuperscript{3}Sun Yat-sen University, China\\[3pt]
\raisebox{-4pt}{\includegraphics[height=16pt]{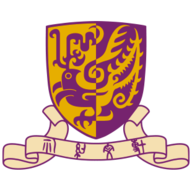}}\hspace{5pt}\textsuperscript{4}The Chinese University of Hong Kong, Hong Kong, China\\[3pt]
\raisebox{-4pt}{\includegraphics[height=16pt]{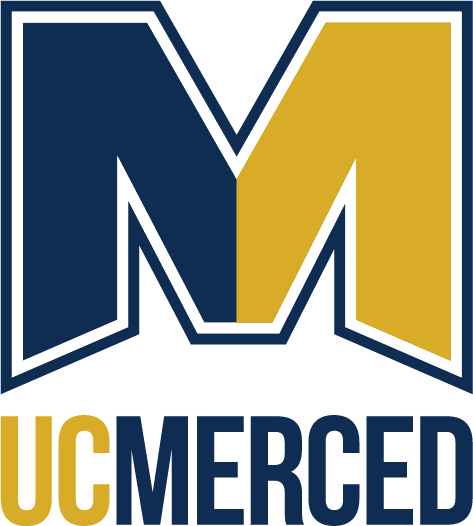}}\hspace{5pt}\textsuperscript{5}University of California, Merced, USA\\[3pt]
\raisebox{-4pt}{\includegraphics[height=16pt]{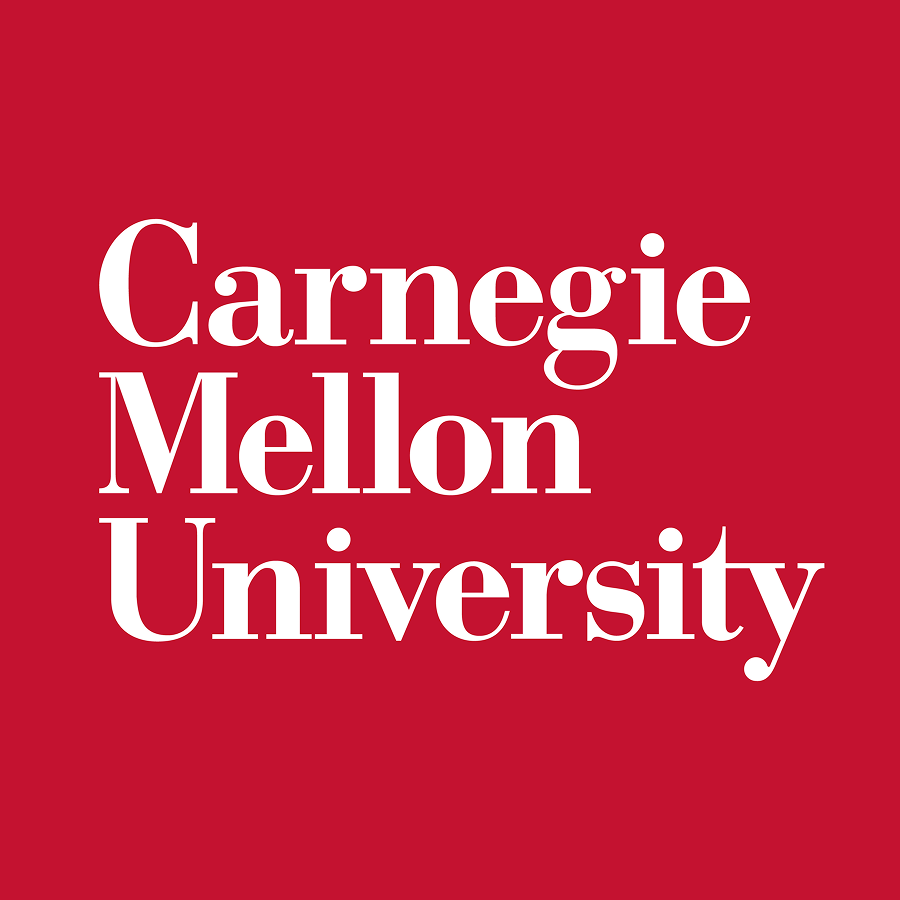}}\hspace{5pt}\textsuperscript{6}Carnegie Mellon University, USA
\end{tabular}}
\newcommand{\reportauthornotes}{%
\textsuperscript{*}Equal contribution.\quad
\textsuperscript{\ensuremath{\dagger}}Corresponding author: \href{mailto:hw5@andrew.cmu.edu}{hw5@andrew.cmu.edu}}

\begin{document}
\begin{reportfrontmatter}

\begin{abstract}
Video generation models synthesize realistic motion, but can they change physical outcomes in response to changes in initial geometry?
We introduce \geoflip{}, a video generation benchmark centered on physical outcome flips induced by geometry.
With shared physical settings and textual prompts held fixed, \geoflip{} applies controlled interventions to support extent, object placement, and mechanism configuration, constructing paired scenes with contrasting reference events, such as remaining stable versus falling or tipping.
These pairs test whether models translate geometric differences in the first frame into correct physical evolution.
The benchmark comprises 482 scene pairs and 964 reference videos across six physical categories, with reference outcomes established through procedural scene construction, physics simulation, and numerical validation and verification of physical events.
We evaluate 16 models for image to video generation along three dimensions: event correctness, structural integrity, and similarity to reference videos.
We examine evaluation consistency with independent vision language judges.
On the 442 conditions shared by all models, the highest event correctness averaged across tasks is 27.98\%, revealing substantial limitations in physical generation conditioned on geometry.
Further analysis shows that model strengths vary across physical mechanisms and that the three evaluation dimensions capture distinct capabilities.
Typical failures include ignoring geometric changes, generating incorrect object motion, bending rigid objects, and introducing additional components.
Rank correlations between judges reach 0.906 for event correctness and 0.929 for structural integrity.
By combining controlled geometric interventions, verifiable physical references, and multidimensional evaluation, \geoflip{} provides a systematic benchmark for assessing progress from visual realism toward physical consistency conditioned on geometry.
\end{abstract}


\end{reportfrontmatter}
\begin{figure}[tbp]
 \centering
 \includegraphics[width=0.82\textwidth]{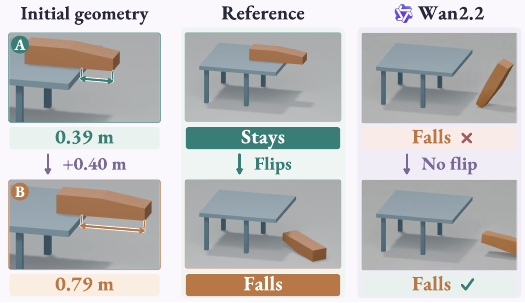}
 \caption{\textbf{Plausible motion can miss geometry-dependent outcomes.} Increasing overhang flips the reference outcome, while Wan2.2-I2V-A14B falls in both conditions. Initial geometry is enlarged (0\,s); outcomes are shown at 1.5\,s.}
 \label{fig:geometry_motivation}
\end{figure}

\section{Introduction}
\label{sec:intro}

A video of a falling object may appear entirely plausible yet be inconsistent with its initial scene~\cite{motamed2026physical}.
For an object resting near a table edge, whether it remains stable or falls depends on its shape, support region, and relative placement; likewise, changing the moment arm of a counterweighted lever can alter its subsequent motion.
Physical consistency in video generation therefore requires both plausible motion and a correct response to the conditions that govern the scene's evolution~\cite{kang2025physical}.
As video generation models increasingly serve to simulate environments and predict future states~\cite{yang2024unisim,wu2025sfp,wu2026futureworlds}, a fundamental question arises: \textbf{\textit{when the initial geometry governing a physical outcome changes, can a model change its generated future accordingly?}}\looseness=-1

Addressing this question requires distinguishing between generating a plausible event and generating the correct event for the given conditions~\cite{zhang2026thinkingvideo}.
Observing a generated video in isolation, even when its motion is natural and its structure coherent, makes it difficult to determine whether the model uses the geometric cues that govern the outcome.
For example, a model biased toward making objects fall off a table may succeed in some scenes while repeating the same motion when the object is adequately supported.
As shown in \cref{fig:geometry_motivation}, increasing the overhang from \textcolor{geometrygreen}{\textbf{0.39\,m}} to \textcolor{geometryorange}{\textbf{0.79\,m}} changes the simulated reference outcome from remaining stable to falling, whereas the illustrated Wan2.2~\cite{wan2025wan22} generations exhibit falling in both conditions.
Success under a single condition therefore does not establish that the model distinguishes the two geometric configurations.
Pairing physically related conditions provides a more direct test of whether generated outcomes change correctly in response to the input differences that determine them.\looseness=-1

Existing benchmarks advance the evaluation of physical video generation from several perspectives.
PhyGenBench, VideoPhy-2, and PhyWorldBench assess adherence to physical laws and commonsense~\cite{meng2025phygenbench,bansal2026videophy2,gu2026phyworldbench}; Physics-IQ compares generated motion against real physical experiments~\cite{motamed2026physical}; and PAI-Bench and PhysInOne use real scenes and simulated data, respectively, to support broader generation, prediction, and understanding tasks~\cite{zhou2026paibench,zhou2026physinone}.
Some also include paired-prompt analyses or motion-transfer pairs~\cite{gu2026phyworldbench,zhou2026physinone}.
However, as summarized in \cref{tab:benchmark_comparison}, testing whether controlled, visible changes to initial geometry under a fixed prompt produce the correct simulation-verified flips in physical events calls for a dedicated evaluation design.\looseness=-1

\begin{table}[!t]
    \centering
    \caption{\textbf{Comparison of benchmark protocols.} \gfyes: supported; \gfpartial: partial; \gfno: absent.}
    \label{tab:benchmark_comparison}
    \footnotesize
    \setlength{\tabcolsep}{2pt}
    \renewcommand{\arraystretch}{1.18}
    \begin{tabularx}{\columnwidth}{@{}ll*{4}{>{\centering\arraybackslash}X}@{}}
        \toprule
        Benchmark & Venue & \shortstack{Ref.\\video} & \shortstack{Physics\\sim.} & \shortstack{Paired\\tests} & \shortstack{Geom.\\flips} \\
        \midrule
        PhyGenBench~\cite{meng2025phygenbench} & ICML 2025 & \gfno & \gfno & \gfno & \gfno \\
        VideoPhy-2~\cite{bansal2026videophy2} & ICLR 2026 & \gfno & \gfno & \gfno & \gfno \\
        PhyWorldBench~\cite{gu2026phyworldbench} & ICLR 2026 & \gfno & \gfno & \gfpartial & \gfno \\
        Physics-IQ~\cite{motamed2026physical} & WACV 2026 & \gfyes & \gfno & \gfno & \gfno \\
        PAI-Bench~\cite{zhou2026paibench} & CVPR 2026 & \gfyes & \gfno & \gfno & \gfno \\
        PhysInOne~\cite{zhou2026physinone} & CVPR 2026 & \gfyes & \gfyes & \gfpartial & \gfno \\
        \midrule
        \rowcolor{comparisonlavender}
        \geoflip{} (core) & --- & \gfyes & \gfyes & \gfyes & \gfyes \\
        \bottomrule
    \end{tabularx}
\end{table}


Motivated by this observation, we introduce \geoflip{}, a video generation benchmark centered on \emph{geometry-induced flips in physical outcomes}.
Building on controlled simulation for physical prediction and counterfactual reasoning~\cite{bear2021physion,yi2020clevrer}, our core tests hold textual prompts and shared physical settings fixed while applying controlled interventions to initial geometry, yielding paired scenes with contrasting reference events.
For example, changing the support position can cause an object to fall rather than remain stable, while shifting a counterweight can release a mechanism that remains balanced under the other condition.
For each condition, the model receives the corresponding first frame and independently generates the subsequent video; text-conditioned models also receive the prompt shared by the pair.
Each intervention specifies a well-defined, observable geometric change whose effect on the target physical event is established by reference simulation.
This design tests whether the model identifies critical geometric differences in the first frame and translates them into the appropriate physical evolution.\looseness=-1

\begin{figure*}[!t]
    \centering
    \captionsetup{skip=4pt}
    \includegraphics[width=\textwidth]{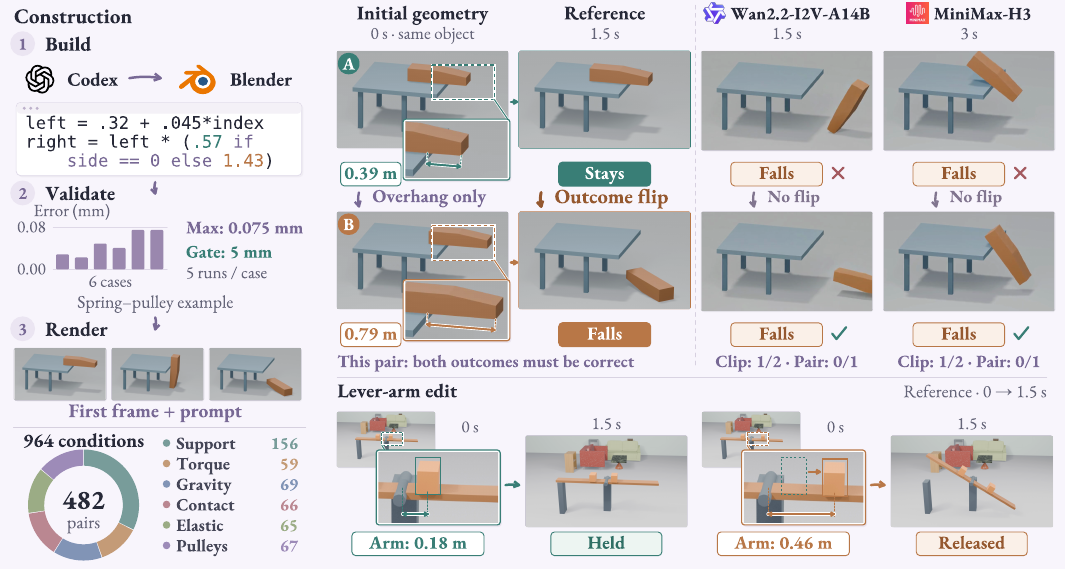}
    \caption{\textbf{Benchmark construction and paired evaluation of geometry-induced physical outcome flips.}}
    \label{fig:benchmark_overview}
\end{figure*}

\Cref{fig:benchmark_overview} illustrates the data construction pipeline and representative paired scenes.
We construct 3D scenes programmatically, simulate their evolution with physics engines, and render first frames and dynamic reference videos in Blender.
Mechanism-specific numerical checks and event-level validation establish reference dynamics and labels, while video checks confirm that key geometric relations and target events remain clearly observable, drawing on the controllability and access to ground truth afforded by synthetic data generation~\cite{greff2022kubric}.
The dataset comprises 482 scene pairs and 964 reference videos across six categories of physical mechanisms: support and static stability, torque-driven rotation, constrained gravitational dynamics, contact and impact, elastic interaction, and pulley-mediated coupling.
These scenes translate geometric responsiveness into observable event differences, including whether an object falls, whether contact occurs, and whether the direction of motion changes.\looseness=-1

For evaluation, we separately assess event correctness, structural integrity, and reference-video similarity, reflecting complementary concerns in video world modeling~\cite{wu2026roboalign,wu2026recap}.
Video-language judges identify generated events and structural violations, and a deterministic scorer compares event predictions with simulation labels.
Structural evaluation examines object bending, additional objects, disappearing components, and changes to supports; PSNR and SSIM provide complementary reference-video similarity measures.
Paired and temporal visualizations further reveal whether models adjust outcomes to geometric changes and whether correct event labels accompany incorrect motion.
The evaluation thus examines both what event occurs and how it unfolds.\looseness=-1

We evaluate 16 image-to-video models and compare event correctness and structural integrity on 442 shared conditions.
Our experiments examine overall performance, differences across physical mechanisms, failure modes, and evaluation reliability.
Event correctness, structural integrity, and reference-video similarity reveal distinct model strengths, and overall leaders do not consistently dominate every physical mechanism.
Case studies further show that models can repeat the same outcome despite a reference-event flip, or generate apparently correct outcomes through erroneous motion, object deformation, or additional components.
On conditions covered by both judges, GPT and Opus achieve Spearman rank correlations above 0.90 for both event and structure scores, indicating strong cross-judge consistency in relative model performance.
Our main contributions are:\looseness=-1
\begin{itemize}[leftmargin=*,itemsep=2pt,topsep=3pt]
    \item[\ding{182}] \textbf{\textit{An evaluation design based on controlled geometric interventions.}}
    We introduce \geoflip{}, which pairs scenes with geometry-induced reference-event flips under fixed prompts and shared physical settings, making geometric responsiveness an explicit, testable target for physical video generation.
    \item[\ding{183}] \textbf{\textit{Validated simulation data and a multidimensional evaluation protocol.}}
    We construct 482 scene pairs spanning six physical categories, establish reference outcomes through numerical and event-level validation, and separately assess event correctness, structural integrity, and reference-video similarity.
    \item[\ding{184}] \textbf{\textit{Systematic experiments and interpretable failure diagnosis.}}
    We evaluate 16 video generation models, reveal mechanism-specific strengths and failures such as correct outcomes with incorrect evolution, and examine model-ranking consistency across judges.
\end{itemize}


\FloatBarrier

\section{Related Work}
\label{sec:related_work}

\begingroup
\makeatletter
\renewcommand{\paragraph}{\@startsection{paragraph}{4}{\z@}{1.25ex plus 0.2ex minus 0.2ex}{-1em}{\normalfont\normalsize\bfseries\itshape}}
\makeatother

\paragraph{Physical consistency in video generation.}
PhyGenBench, VideoPhy-2, and PhyWorldBench evaluate generated videos in terms of physical laws and commonsense~\cite{meng2025phygenbench,bansal2026videophy2,gu2026phyworldbench}.
Physics-IQ uses real physical experiments as references to assess models' ability to predict physical processes~\cite{motamed2026physical}, while PAI-Bench and PhysInOne cover broader physical generation, prediction, and understanding tasks~\cite{zhou2026paibench,zhou2026physinone}.
Complementary work studies physically informed prediction and dynamics generalization~\cite{wu2024pastnet,wu2024prometheus,wu2024pure}.
\mbox{\geoflip{}} further focuses on how generated outcomes respond to initial geometry: with textual prompts and shared physical settings fixed, can a model generate distinct, correct physical events as the geometric configuration changes?\looseness=-1

\paragraph{Physical understanding and paired evaluation.}
Physion and CLEVRER use controlled physical scenes to study event prediction and reasoning~\cite{bear2021physion,yi2020clevrer}.
MVP reduces shortcut reliance by pairing visually similar videos with identical questions but different correct answers~\cite{krojer2025mvp}.
Its paired score requires both answers to be correct, exposing distinctions that success on individual examples can obscure.
\mbox{\geoflip{}} adopts the principle of jointly evaluating related conditions, but shifts the task from recognizing events in provided videos to generating futures from initial frames.
We assess generated events against their condition-specific references and use paired comparisons to expose failures to respond to geometric changes.\looseness=-1

\paragraph{Controlled interventions and counterfactual video generation.}
What-If World changes physical details in prompts to test whether generated outcomes exhibit the expected differences~\cite{cai2026whatifworld}.
CausalCascade uses simulated scenes and controlled counterfactual variants to examine causal event cascades and responses to physical interventions~\cite{li2026causalcascade}.
The core tests in \mbox{\geoflip{}} hold textual prompts and shared physical settings fixed, encode interventions as visible geometric changes in the initial frame, and establish event flips through reference simulation.
Evaluation requires each generated event to match its respective reference, rather than merely requiring different outcomes across conditions.
Causal prediction also studies interventions on contextual regions~\cite{wang2024nuwadynamics,duan2024capaint}.
\par
\endgroup
\looseness=-1

\begin{figure*}[!t]
    \centering
    \captionsetup{skip=4pt}
    \includegraphics[width=\textwidth]{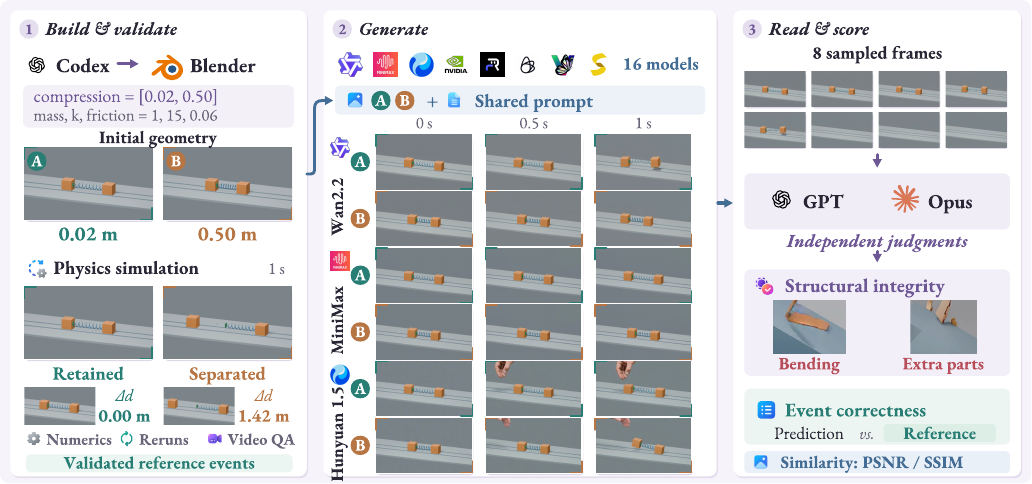}
    \caption{\textbf{Construction and evaluation protocol.} Controlled geometry edits yield validated reference events; $\Delta d$ denotes separation change over 0--1 s. Wan2.2, MiniMax, and HunyuanVideo-1.5 independently generate from each first frame and a shared prompt. GPT and Opus read sampled frames; reference labels enter only the event scorer. Insets illustrate structural violations. PSNR/SSIM assess reference similarity.}
    \label{fig:method_protocol}
\end{figure*}


\section{The \geoflip{} Benchmark}
\label{sec:benchmark}

\geoflip{} evaluates whether video generation models identify critical geometric differences in the initial scene and generate the corresponding physical events.
As illustrated in Fig.~\ref{fig:method_protocol}, the benchmark combines controlled pair construction, reference simulation and validation, and generated-video evaluation.
Geometric interventions produce paired scenes with different reference events; event correctness, structural integrity, and reference similarity characterize the generated results.

\subsection{Controlled Pair Construction}
\label{sec:paired_construction}

Each core sample consists of two scenes with shared physical settings.
We fix common parameters, such as mass and friction, and the textual prompt, while changing a specified aspect of the initial geometry to produce different reference events.
For example, increasing initial spring compression from 0.02 m to 0.50 m changes the reference outcome from retained contact to separation between the block and the piston (Fig.~\ref{fig:method_protocol}).
For the $i$-th pair, construction and generation take the form
\begin{equation}
\begin{aligned}
 s_i^A &= (g_i^A,\theta_i),\quad s_i^B=(g_i^B,\theta_i),\\
 y_i^A &\ne y_i^B,\quad
 \hat V_i^c\sim G(\,\cdot\mid I_i^c,p_i),
\end{aligned}
\label{eq:paired_generation}
\end{equation}
where $c\in\{A,B\}$, $g_i^c$ denotes the initial geometry, $\theta_i$ denotes the shared physical parameters, and $y_i^c$ denotes the reference event within a predefined observation window.
The model $G$ receives the corresponding first frame $I_i^c$ and shared prompt $p_i$, and independently generates each video $\hat V_i^c$.

Each intervention specifies an observable geometric change whose physical outcome is established by simulation.
Comparing generated events across the two conditions tests whether the model translates the input difference into the corresponding change in physical outcome.
Auxiliary controls are recorded separately from the core outcome-flip pairs.

\subsection{Reference Simulation and Validation}
\label{sec:reference_validation}

We construct 3D scenes through programmatic scripts, simulate motion with physics engines, and render first frames and reference videos in Blender.
Each condition retains scene parameters, motion trajectories, event labels, and validation records, linking its visible configuration to the resulting physical evolution.

Reference validation combines numerical and video checks.
For each physical mechanism, numerical checks use analytical or independent numerical references within their applicable intervals, together with time-step refinement, unit rescaling, and independent reruns to assess trajectory errors and event consistency.
Contact and collision checks additionally examine convergence, penetration, and nonphysical energy increases.
Video checks confirm that key objects, contact relations, and target events remain clearly observable after rendering.
Figure~\ref{fig:method_protocol} shows validated reference outcomes at 1 s alongside independently generated video sequences.
A pair enters the core evaluation when both conditions pass validation and their reference events differ.
Mechanism-specific configurations and thresholds appear in the supplementary material.

\subsection{Evaluation Protocol}
\label{sec:event_evaluation}

We separately assess whether the generated event is correct, whether the scene remains structurally intact, and how closely the video resembles its reference.
GPT and Opus independently read sampled frames and return an event label and a structural-violation label.
A deterministic scorer compares the predicted event with the simulation label.
Structural labels capture object bending, extra objects, disappearing components, and changes to supports.

\noindent\textbf{\textit{Event correctness and structural integrity.}}
Event correctness uses balanced accuracy within each task, followed by an equal-weight average across tasks.
Structural integrity measures the proportion of conditions explicitly labeled \texttt{clear}, also averaged equally across tasks:
\begin{equation}
\begin{aligned}
E_j &= \frac{1}{|\mathcal T|}\sum_{t\in\mathcal T}
       \frac{1}{|\mathcal C_t|}\sum_{k\in\mathcal C_t} R_{j,t,k},\\[2pt]
S_j &= \frac{1}{|\mathcal T|}\sum_{t\in\mathcal T}
       \frac{\sum_{n\in\mathcal D_t}\mathbf{1}[\hat z_{j,n}=\mathrm{clear}]}{|\mathcal D_t|}.
\end{aligned}
\label{eq:event_structure_scores}
\end{equation}
Here, $j$ indexes the judge, $\mathcal T$ is the task set, $\mathcal C_t$ is the set of reference event classes for task $t$, and $\mathcal D_t$ contains its evaluated conditions.
$R_{j,t,k}$ is recall for class $k$: the fraction of conditions with reference label $k$ that judge $j$ correctly labels as $k$.
The structural label is $\hat z_{j,n}$.
Unrecognized or uncertain events count as incorrect; flagged and uncertain structural judgments remain in the denominator of $S_j$.
We report event and structure scores separately, distinguishing a correct target event from structural failures during its execution.
For example, a video can depict the expected fall while also bending the supporting table.

\noindent\textbf{\textit{Reference similarity.}}
PSNR and SSIM measure pixel-level and image-structure similarity to reference videos, complementing the event and structure assessments.
We retain each judge's individual scores; sampling configurations and aggregate ranking are specified in the experimental setup.


\FloatBarrier


\begin{table*}[!t]
\centering
\caption{\textbf{Overall model ranking.} Event and structure scores are task-macro percentages on 442 shared \geoflip{} conditions, judged by GPT. PSNR/SSIM use 40-clip. Rows follow the mean rank across the three evaluation dimensions; colors mark the top three models.}
\label{tab:rq1_overall}
\includegraphics[width=\textwidth]{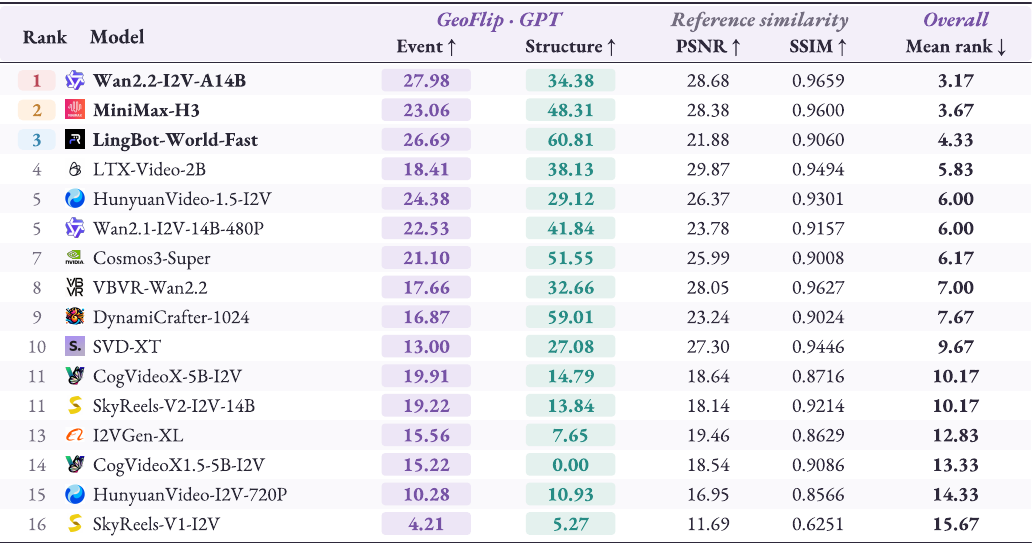}
\end{table*}


\section{Experiments}
\label{sec:experiments}

Our experiments address four research questions:
\begin{itemize}[leftmargin=*,itemsep=1pt,topsep=2pt]
    \item[\ding{182}] \textbf{\textit{RQ1: Overall performance and evaluation dimensions.}}
    How do video generation models perform in event correctness, structural integrity, and reference-video similarity, and do these dimensions reveal consistent model strengths?
    \item[\ding{183}] \textbf{\textit{RQ2: Differences across physical mechanisms.}}
    How do event correctness and structural integrity vary across physical mechanisms, and do overall leaders retain their advantages across categories?
    \item[\ding{184}] \textbf{\textit{RQ3: Physical failure modes.}}
    What typical failures occur in responding to geometric changes, generating motion, and preserving object structure?
    \item[\ding{185}] \textbf{\textit{RQ4: Evaluation reliability.}}
    How accurately do automated judges recognize reference physical events, and how consistent are their judgments of generated videos and model rankings?
\end{itemize}

\subsection{Experimental Setup}
\label{sec:experimental_setup}

\noindent\ding{118}\ \textbf{\textit{Evaluation data.}}
The selected \geoflip{} evaluation pool contains 720 conditions across 43 tasks.
The main comparison uses the intersection of available GPT event judgments across all 16 models: 442 conditions spanning 34 tasks.
Event and structure scores use exactly the same conditions for every model.
Paired qualitative comparisons follow the geometric-intervention criteria in Sec.~\ref{sec:paired_construction}.

\vspace{1pt}
\noindent\ding{118}\ \textbf{\textit{Models and inference.}}
We evaluate 16 model variants in the image-to-video setting:
LTX-Video-2B~\cite{hacohen2024ltxvideo},
SVD-XT~\cite{blattmann2023svd},
CogVideoX-5B-I2V~\cite{yang2025cogvideox},
Wan2.1-I2V-14B-480P~\cite{wan2025wan},
HunyuanVideo-I2V-720P~\cite{kong2024hunyuanvideo},
SkyReels-V1-I2V~\cite{skyreels2025v1},
I2VGen-XL~\cite{zhang2023i2vgen},
DynamiCrafter-1024~\cite{xing2024dynamicrafter},
SkyReels-V2-I2V-14B~\cite{chen2025skyreelsv2},
Wan2.2-I2V-A14B~\cite{wan2025wan22},
CogVideoX1.5-5B-I2V~\cite{yang2025cogvideox},
MiniMax-H3~\cite{minimax2026h3},
VBVR-Wan2.2~\cite{wang2026vbvr},
HunyuanVideo-1.5-I2V~\cite{tencent2025hunyuan15},
Cosmos3-Super~\cite{nvidia2026cosmos3},
and LingBot-World-Fast~\cite{robbyant2026lingbot}.
Each model generates independently from the corresponding first frame; text-conditioned models also receive the shared prompt for each condition.
We retain model-specific inference configurations and record resolution, frame rate, output length, sampling steps, and guidance settings.
Time-specific comparisons require outputs that cover the prescribed observation time; video duration is distinct from event correctness.

\vspace{1pt}
\noindent\ding{118}\ \textbf{\textit{Event reading.}}
The main results use GPT-6-Astra judgments from eight sampled frames per video.
Each event judgment also returns a separate structural-violation label, from which we compute structural integrity.
Opus judgments on the earlier 346-condition subset support cross-judge comparisons separately from the main ranking.

\vspace{1pt}
\noindent\ding{118}\ \textbf{\textit{Metrics.}}
Following Sec.~\ref{sec:event_evaluation}, we compute task-macro event correctness and structural integrity over the shared 442 conditions and 34 tasks.
The overall score averages three ranks: event correctness, structural integrity, and reference similarity, where the last is the mean of PSNR and SSIM ranks.
The three dimensions receive equal weight; component ties receive average ranks, and a lower overall score is better. Models with identical overall scores share a rank.\par

\vspace{1pt}
\noindent\ding{118}\ \textbf{\textit{Reference similarity.}}
We use PSNR and SSIM from the released 40-clip evaluation reports, retaining their original aggregation protocol. All 16 models cover the same 40 reference clips; PSNR is reported in dB.\par

\subsection{Overall Performance (RQ1)}
\label{sec:rq1_results}

\begin{figure}[t]
\centering
\includegraphics[width=\columnwidth]{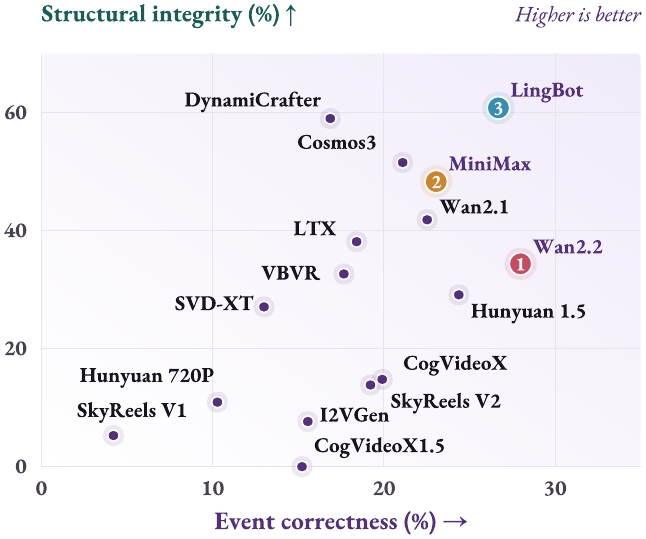}
\caption{\textbf{Event correctness and structural integrity.} Each point represents one model on the same 442 conditions. Both axes report task-macro percentages from GPT judgments. Numbered markers denote the top three overall ranks in Table~\ref{tab:rq1_overall}.}
\label{fig:rq1_event_structure}
\end{figure}

Table~\ref{tab:rq1_overall} presents the overall ranking across event correctness, structural integrity, and reference similarity.
Wan2.2, MiniMax-H3, and LingBot-World-Fast rank first, second, and third, with mean ranks of 3.17, 3.67, and 4.33, respectively.
Their strengths differ: Wan2.2 achieves the highest event correctness at 27.98\%, LingBot achieves the highest structural integrity at 60.81\%, and MiniMax combines 48.31\% structural integrity with strong reference similarity.
Figure~\ref{fig:rq1_event_structure} further illustrates the difference between event and structure performance: DynamiCrafter achieves 59.01\% structural integrity but only 16.87\% event correctness, whereas Wan2.2 leads in event correctness with 34.38\% structural integrity.
These results show that generating the correct physical event and maintaining the apparatus during its evolution are distinct capabilities that require separate evaluation.

Reference similarity provides a complementary evaluation perspective.
LTX achieves the highest PSNR at 29.87 dB, but its event correctness is 18.41\%.
LingBot has a lower PSNR of 21.88 dB, yet achieves 26.69\% event correctness and 60.81\% structural integrity.
Wan2.2 leads in both SSIM (0.9659) and event correctness, but its structural integrity trails MiniMax and LingBot.
The overall ranking summarizes performance across the three dimensions, while the individual scores reveal differences between visual fidelity, event accuracy, and structural preservation.

\begin{figure*}[t]
\centering
\includegraphics[width=\textwidth]{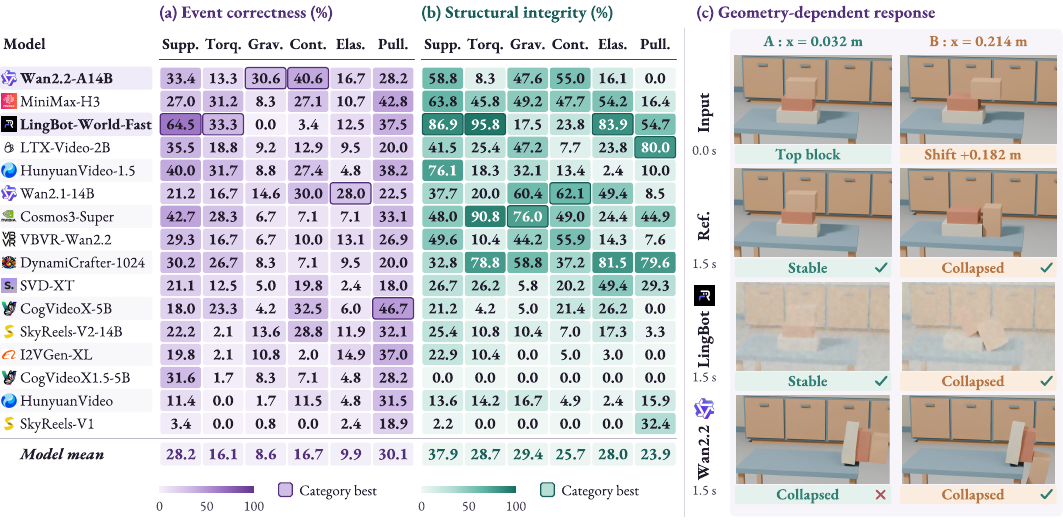}
\caption{\textbf{Performance across physical mechanisms.} (a,b) Task-macro scores within each category; outlined cells mark category leaders. Supp., Torq., Grav., Cont., Elas., and Pull. denote support, torque, gravity, contact, elastic, and pulley mechanisms. (c) Shifting the top block by 0.182 m changes the reference from stable to collapsed. LingBot reproduces this contrast; Wan2.2 collapses in both conditions. Outcomes are shown at 1.5 s.}
\label{fig:rq2_categories}
\end{figure*}

\subsection{Performance across Physical Mechanisms (RQ2)}
\label{sec:rq2_results}

Figure~\ref{fig:rq2_categories}(a,b) decomposes the same 442 conditions into six physical categories, averaging task scores within each category.
Across the 16 models, mean event correctness is highest for pulley coupling (30.1\%) and support stability (28.2\%), but falls to 8.6\% for constrained gravitational dynamics and 9.9\% for elastic interaction.
Structural integrity follows a different pattern: support stability achieves the highest mean (37.9\%), whereas pulley coupling has the lowest (23.9\%) despite its higher event correctness.
Correct event prediction therefore does not ensure that the underlying mechanism remains structurally intact.

Model strengths also vary by mechanism.
Wan2.2 leads event correctness in constrained gravitational dynamics (30.6\%) and contact and impact (40.6\%), while LingBot leads in support stability (64.5\%) and torque-driven rotation (33.3\%); Wan2.1 leads in elastic interaction (28.0\%).
The paired example in Fig.~\ref{fig:rq2_categories}(c) illustrates LingBot's advantage in support stability: moving the top block by 0.182 m changes the reference from stable to collapsed, and LingBot reproduces both outcomes.
Wan2.2 instead collapses the stack in both conditions, despite producing clearer images.
Thus, overall leadership does not imply uniformly stronger geometric responses across physical mechanisms.

\begin{figure*}[t]
\centering
\includegraphics[width=\textwidth]{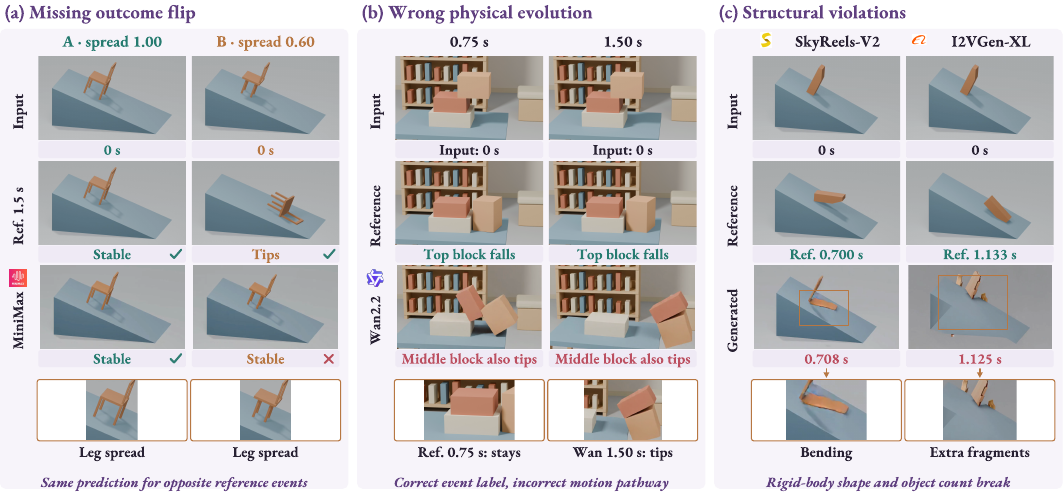}
\caption{\textbf{Three types of physical failure.} (a) MiniMax-H3 misses the geometry-induced outcome flip. (b) Wan2.2 produces the correct collapse label but also overturns the middle block. (c) SkyReels-V2 bends a rigid object, while I2VGen-XL creates extra fragments. Insets highlight geometry and failure details; timestamps identify the displayed frames.}
\label{fig:rq3_failures}
\end{figure*}

\begin{figure}[!t]
\centering
\includegraphics[width=\columnwidth]{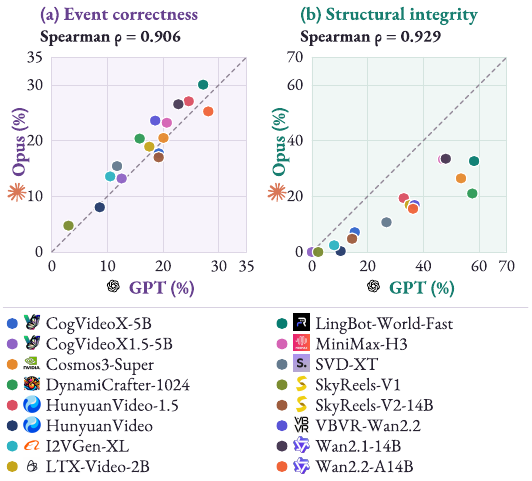}
\caption{\textbf{Cross-judge score consistency.} GPT and Opus assess the same 346 conditions for 16 models. Scores are task-macro percentages over 30 tasks; structural integrity is the clear rate. Colors identify models consistently across both panels, dashed lines indicate equal scores, and Spearman $\rho$ measures ranking consistency.}
\label{fig:rq4_judge_consistency}
\end{figure}

\subsection{Failure Mode Analysis (RQ3)}
\label{sec:rq3_results}

Figure~\ref{fig:rq3_failures} presents three representative failure modes.
First, models fail to adjust generated events to changes in initial geometry.
In Fig.~\ref{fig:rq3_failures}(a), reducing the chair's leg spread changes the reference outcome from stable to tipping, whereas MiniMax-H3 keeps the chair upright in both conditions.
Second, a correct event label does not imply a correct physical evolution.
In Fig.~\ref{fig:rq3_failures}(b), only the top block collapses in the reference simulation, while the middle and bottom blocks remain stable.
Wan2.2 also generates a collapse, but overturns the middle block as well, changing which objects move and how they interact through contact.

Structural violations further undermine physical consistency.
In Fig.~\ref{fig:rq3_failures}(c), SkyReels-V2 visibly bends a rigid object, while I2VGen-XL creates additional fragments absent from the initial scene.
These failures motivate joint event and structure evaluation, supported by temporal visualizations that expose process errors beyond outcome labels.

The paired construction makes these failures interpretable.
MiniMax repeats a stable outcome when the reference changes, exposing insensitivity to the intervention.
Wan2.2 produces the correct event category through an incorrect set of moving objects, while SkyReels and I2VGen alter the objects themselves.
These comparisons distinguish failures in responding to geometric conditions, determining which objects move, and preserving the physical mechanism.

\begin{table}[t]
\centering
\caption{\textbf{Evaluation reliability.} Reference-event recognition by Opus and cross-judge agreement on the shared evaluation subset.}
\label{tab:rq4_reliability}
\includegraphics[width=\columnwidth]{assets/asset09_rq4_reliability_table.pdf}
\end{table}

\subsection{Evaluation Reliability and Cross-Judge Stability (RQ4)}
\label{sec:rq4_results}

We examine evaluation reliability through reference-event recognition and cross-judge consistency (Table~\ref{tab:rq4_reliability}).
On 346 simulation reference videos with known event labels, Opus achieves 94.22\% event recognition accuracy and 93.72\% task-macro balanced accuracy, demonstrating accurate recognition of most reference events.
On the 346 conditions jointly covered by both judges across 16 models, GPT and Opus agree on 81.54\% of event labels and 73.55\% of structural labels across 5,536 generated videos.
The lower structural agreement indicates greater disagreement in identifying structural violations.

Figure~\ref{fig:rq4_judge_consistency} compares the model scores assigned by the two judges.
Spearman rank correlations reach 0.906 for event correctness and 0.929 for structural integrity, indicating strong cross-judge stability in relative model performance.

The magnitude of disagreement further distinguishes the two evaluation dimensions.
Across the 16 models, the mean absolute score difference between GPT and Opus is 2.47 percentage points for event correctness, compared with 15.14 points for structural integrity.
For DynamiCrafter, structural integrity changes from 57.64\% under GPT to 21.03\% under Opus, a gap of 36.61 points.
All nonzero structural score differences favor GPT.
Together with the high rank correlation, this pattern shows that the judges largely retain the ordering of structural quality while applying different levels of strictness to the same generated videos.
The larger structural score gaps identify structural assessment as the main source of judge sensitivity, motivating explicit checks of object shape and component persistence alongside recognition of the final physical event.


\FloatBarrier

\section{Conclusion}
\label{sec:conclusion}

We introduce \geoflip{}, a video generation benchmark centered on geometry-induced physical outcome flips.
Through controlled scene pairs, validated simulation references, and separate event and structure evaluation, \geoflip{} tests whether models translate initial geometric differences into correct physical evolution.
Evaluation of 16 video generation models reveals distinct strengths across physical mechanisms and differences between visual quality, event correctness, and structural integrity.
Even correct outcome labels can accompany erroneous motion, object deformation, or additional components.
Cross-judge experiments further show strong consistency in relative model rankings.
\geoflip{} provides a controlled, interpretable basis for evaluating geometry-conditioned video generation and identifies key challenges in progressing from visual plausibility to physical consistency.


\end{document}